\pdfoutput=1

\documentclass[11pt]{article}

\usepackage[]{acl}

\usepackage{times}
\usepackage{latexsym}

\usepackage[T1]{fontenc}

\usepackage[utf8]{inputenc}

\usepackage{microtype}

\usepackage[utf8]{inputenc}
\usepackage{graphicx}
\graphicspath{ {./} }
\usepackage{amsmath}
\usepackage{amsthm}
\usepackage{booktabs}
\usepackage[ruled,vlined]{algorithm2e}
\usepackage{algorithmic}
\usepackage{color}
\usepackage{multirow}
\usepackage{newtxmath}
\usepackage{bm}
\usepackage{arydshln}
\usepackage{marvosym}
\usepackage{subfig}
\title{A Coarse-to-fine Cascaded  Evidence-Distillation Neural Network for Explainable Fake News Detection}

\author{
	Zhiwei Yang$^{1,2,3,5}$, Jing Ma$^{2,*}$, Hechang Chen$^{3,5,*}$, Hongzhan Lin$^{2}$, Ziyang Luo$^{2}$, Yi Chang$^{3,4,5,}\thanks{{\; Corresponding authors. }}$ \\
	$^1$ College of Computer Science and Technology, Jilin University, Changchun, China\\
	$^2$ Department of Computer Science, Hong Kong Baptist University, Hong Kong, China\\
	$^3$ School of Artificial Intelligence, 
	$^4$ International Center of Future Science, Jilin University, China \\
	$^5$ Key Laboratory of Symbolic Computation and Knowledge Engineering of Ministry of Education\\
    \texttt{yangzw18@mails.jlu.edu.cn, chenhc@jlu.edu.cn,  }\\
    \texttt{\{majing, cszyluo, cshzlin\}@comp.hkbu.edu.hk, yichang@jlu.edu.cn}
}

\begin{document}
\maketitle

\begin{abstract}
Existing fake news detection methods aim to classify a piece of news as true or false and provide veracity explanations, achieving remarkable performances. However, they often tailor automated solutions on manual fact-checked reports, suffering from limited news coverage and debunking delays. When a piece of news has not yet been fact-checked or debunked, certain amounts of relevant raw reports are usually disseminated on various media outlets, containing the wisdom of crowds to verify the news claim and explain its verdict. 
In this paper, we propose a novel Coarse-to-fine Cascaded Evidence-Distillation (CofCED) neural network for explainable fake news detection based on such raw reports, alleviating the dependency on fact-checked ones. Specifically, we first utilize a hierarchical encoder for web text representation, and then develop two cascaded selectors to select the most explainable sentences for verdicts on top of the selected top-$K$ reports in a coarse-to-fine manner.  
Besides, we construct two explainable fake news datasets, which are publicly available. Experimental results demonstrate that our model significantly outperforms state-of-the-art baselines and generates high-quality explanations from diverse evaluation perspectives. 
\end{abstract}

\section{Introduction}
During the COVID-19 pandemic, almost 80\% of consumers in the United States received fake news, which has caused confusion and undermined public health efforts\footnote{ \url{https://www.statista.com/topics/3251/fake-news}}. 
The proliferation of fake news has increased the demand for automatic fake news detection~\cite{guo2022survey}. 
To further clarify and explain detection results, 
explainable fake news detection has gained more importance recently, 
aiming to classify the truthfulness of a piece of news and generate veracity explanations\footnote{Explanations and evidence are used interchangeably}~\cite{kotonya2020survey}. 
However, 
existing methods have a limitation in detecting fake news timely as they heavily relied on debunked reports of investigated journalism. Thus, it is urgent to develop explainable yet general methods to mitigate this issue. 

\begin{figure} [t!]
	\centering
	\includegraphics[width=0.98 \linewidth]{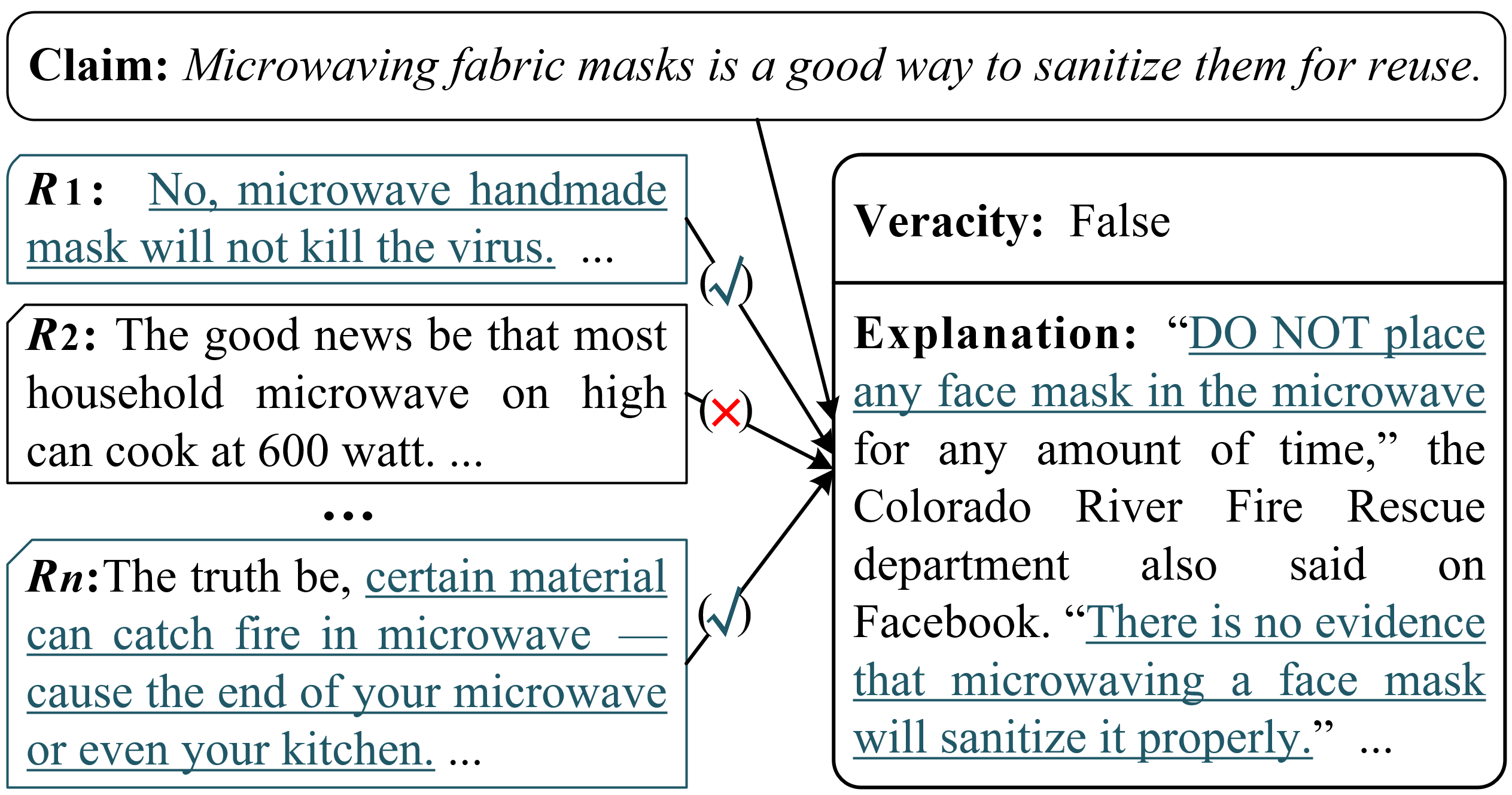} 
	\caption{
    An example for veracity explanation generation.   
	The underlined explanations can be semantically inferred from some relevant sentences in the reports $R_1$ and $R_n$.  ``\textbf{$R$}'' denotes the raw report. 
    }
    \label{demo}\vspace{-0.4cm}
\end{figure}
Many previous approaches detected fake news without any justifications~\cite{wang2017liar,ma2018detect}. 
Recently, 
some explainable methods highlighted salient words or phrases in relevant reports as explanations~\cite{popat2018declare,wu2021evidence}, which lack readable complete sentences. 
To alleviate these issues, some methods aimed to extract salient sentences from relevant reports via attention mechanisms~\cite{nie2019combining,ma2019sentence}, or pre-trained extractive-abstractive summarization~\cite{kotonya2020health}, etc. 
As the human justification about veracity labels can significantly improve the performance of veracity prediction~\cite{alhindi2018liarplus}, 
\citet{atanasova2020generating} proposed the first study on producing veracity explanations jointly with veracity prediction utilizing the debunked report released by fact-checking websites. 
%
However, such a debunked report is based on manual endeavors, thus prone to be coverage-limited and relatively inefficient. 

A new study by MIT researchers suggests that crowds of laypeople reliably rate claims as effectively as fact-checkers do ~\cite{allen2021scaling}. To use the wisdom of crowds, 
we assume that crowds of relevant raw reports (e.g., media reports, user comments, blogs, etc.) published by different media outlets contain evidence for effectively detecting fake news and explaining verdicts ~\cite{ma2019sentence,popat2018declare}. 
As shown in Figure~\ref{demo}, given a false claim ``\textit{Microwaving fabric masks is a good way to sanitize them for reuse}'', the check-worthy reports $R_1$ and $R_n$ are selected from all reports $[R_1, R_2, \cdots, R_n]$ and then some evidential sentences (underlined) can be used to generate veracity explanations. In contrast, existing methods usually tailor models on one manual fact-checked article, rarely attempting to detect fake news based on raw reports. 

To this end, we propose a general coarse-to-fine cascaded evidence-distillation (CofCED) network to detect fake news and explain verdicts directly using raw reports, mitigating the dependency on fact-checked reports.  
Specifically, we design a hierarchical encoder for text representation, 
and then we develop two coarse-to-fine cascaded selectors to distill explainable sentences on top of the selected top-$K$ check-worthy reports. 
Our predictions of explainable sentences can be obtained by explicitly considering four features, i.e., claim relevance, richness, salience, and non-redundancy. 
Different from FEVER~\cite{thorne2018fever} using 
human-crafted claims with credible Wikipedia articles, the claims in our task are real-world news containing some unreliable reports. 
Thus, detecting fake news on raw reports is much more challenging and significant than that in FEVER task. 

%
Our contributions are  as follows: \textbf{1) } To the best of our knowledge, we present the first study on explainable fake news detection directly utilizing the wisdom of crowds, alleviating the dependency on fact-checked reports; \textbf{2)} Our model has the advantage of revealing insight into the generation of veracity explanations from various perspectives;  
\textbf{3)} We construct two realistic datasets, i.e., RAWFC and LIAR-RAW, consisting of raw reports for each claim. Experimental results on benchmarks demonstrate the effectiveness of CofCED for detecting fake news and and explaining verdicts based on raw reports. 
Our resources are publicly available at 
\href{https://github.com/Nicozwy/CofCED}{\color{blue}{https://github.com/Nicozwy/CofCED}}. 


\section{Related Work}
We review prior works closely related to ours based on several  surveys \cite{shu2017fake,kotonya2020survey}.

\textbf{Black-boxed fake news detection.} 
Many existing studies on fake news detection achieved promising performances by incorporating claim metadata to facilitate the detection, such as user profiles~\cite{wang2017liar,long2017fake,karimi2018multi}. 
Besides, 
various deep learning methods have been proposed to capture report features, e.g., credibility \cite{popat2017cnn}, stances \cite{ma2018detect}, writing styles \cite{potthast2018stylometric}, 
extra   knowledge~\cite{dun2021kan}, etc. Although these methods could improve the detection performance, they are lack of explainability on verdicts. 

\textbf{Explainable fake news detection.}
To address the above issue, 
many explainable methods on this task explored attention mechanisms to highlight salient words~\cite{popat2018declare,wu2021evidence}, news attributes~\cite{yang2019xfake}, and suspicious users~\cite{lu2020gcan}, to obtain relevant evidence, providing a certain explainability. 
To improve the readability in word-level methods, there are some methods obtained evidential sentences using attention weights~\cite{shu2019defend}, semantic matching~\cite{nie2019combining}, and entailment~\cite{ma2019sentence}.  
More recently, Atanasova et al.~\shortcite{atanasova2020generating} proposed the first study on directly producing veracity explanations using extractive summarization, and Kotonya and Toni~\shortcite{kotonya2020health} made use of  extractive-abstractive summarization for explanation generation, independent of the veracity prediction. However, they significantly relied on the manual fact-checked report and rarely attempted to consider fine-grained features for this task. 
Thus, we utilize the wisdom of crowds for fake news detection based on raw reports, providing a highly explainable structure for explanation generation. 


\textbf{Datasets}. For explainable fake news detection, FEVER~\cite{thorne2018fever} was crafted merely from credible Wikipedia articles, and  MultiFC~\cite{augenstein2019multifc} provided a real-world benchmark for multi-domain claims. While offering evidence labels, they do not contain veracity explanations. By contrast, LIAR-PLUS~\cite{alhindi2018liarplus} extended on LIAR~\cite{wang2017liar} and PUBHEALTH ~\cite{kotonya2020health} on the public health, providing manual explanations for explainable fake news detection. 
However, they only contain the manual fact-checked report that is relatively inefficient and coverage-limited. Thus, we constructed two datasets by collecting raw reports, which is more suitable and challenging for this task. 

\begin{figure*} [t!]
	\centering
	\includegraphics[width=1 \textwidth]{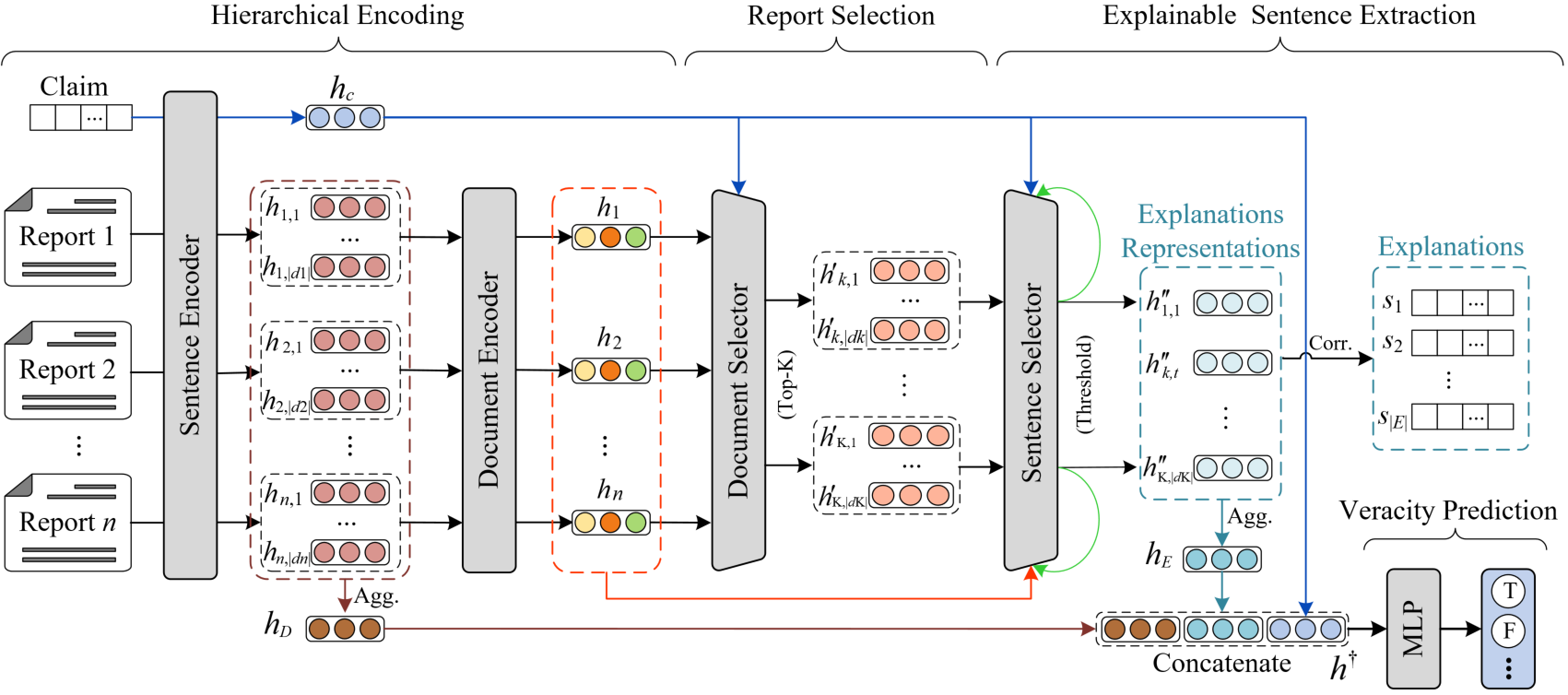} 
	\caption{
	An overview of our proposed CofCED framework. The document selector and the sentence selector are used for selecting check-worthy reports (containing oracles) and oracles, respectively. ``Agg.'' denotes aggregation and `` Corr.'' denotes corresponding. We use different color to highlight different objects. Note that the green line denotes the last output of sentence selection for checking redundancy. 
    }
    \label{framework} 
\end{figure*}
\section{Problem Statement}
Given a fake news dataset $\{\mathcal{C}\}$, $\mathcal{C} = (c, \mathcal{D})$ is a tuple representing a given claim $c$ and its relevant raw reports $\mathcal{D}=\{d_i\}_{i=1}^{|\mathcal{D}|}$,  where each $d_i=(s_{i,1},s_{i,2},\cdots,s_{i,|d_i|})$ denotes a relevant report consisted of a sequence of sentences and $|.|$ denotes the number of items.  
In the task of explainable fake news detection, each claim $c$ is associated with a veracity $y$ taking one of the class labels from $\{\text{True}, \text{False}, \cdots \}$, 
and each raw report $d_i$ is associated with a binary label $y_i^{d} \in Y^{d}$ indicating that whether $d_i$ contains explainable sentences (i.e., oracles). 
For each sentence $s_{i,j}$, $y_{i,j}^s \in Y^s$ is a binary label indicating that whether $s_{i,j}$ is one of the explainable sentences w.r.t. the gold justification. 

We formulate this task as a multi-task learning problem by considering check-worthy report selection, explainable sentence extraction, and veracity prediction.  
Formally, $f: f(c, \mathcal{D}) \rightarrow (\hat{y}, \hat{Y}^{d}, \hat{Y}^s, \hat{E})$, 
where $\hat{E}$ denotes the veracity explanation (i.e., \textit{evidence}) consisting of a set of predicted sentences (i.e.,  $\hat{y}_i^{d}=1$ and $\hat{y}_{i,j}^s=1$). 
\section{CofCED: The Proposed Method}

Fig. \ref{framework} gives an overview of our proposed CofCED, which consists of four parts: hierarchical encoding, report selection, explainable sentence extraction, and veracity prediction.

\subsection{Hierarchical Encoding}
Given a word sequence of a claim or report sentence $T = (w_1 \cdots w_t \cdots w_{|T|})$, where $w_t \in \vmathbb{R}^d$ is a $d$-dimensional vector initialized with a text encoder. Because words form a sentence and sentences form a report, we utilize a hierarchical encoding method for sentence and report representation in our model. Specifically, 
for sentence encoding, we use the special token ``[CLS]'' embedding from the final contextual layer of the pre-trained language model \cite{sanh2019distilbert} as the sentence representation. 
Thus, we obtain the sentence representation for a claim $c$ and each sentence $s_{i,j}$ in a raw report $d_i$ as $\mathbf{h}_c \in \vmathbb{R}^d$ and $\mathbf{h}_{i,j} \in \vmathbb{R}^d$, respectively. 

For document encoding, we further adopt a document encoder consisting of a bidirectional LSTM (BiLSTM) \cite{rashkin2017lstm} and a max-pooling layer to aggregate all salient sentence features as the representation of a report:  
\begin{flalign}
\mathbf{\tilde{h}}_{i,j} &=\text{BiLSTM}(\mathbf{h}_{i,j}, \overrightarrow{\mathbf{h}}_{i,j-1}, \overleftarrow{\mathbf{h}}_{i,j-1}, \theta) \\
\mathbf{h}_{i} &= \text{Max} ([\mathbf{\tilde{h}}_{i,1}; \mathbf{\tilde{h}}_{i,2}; \cdots; \mathbf{\tilde{h}}_{i,|d_i|}]) 
\label{eq:lstm}
\end{flalign}
where $\mathbf{\tilde{h}}_{i,j} \in \vmathbb{R}^{d} $ denotes the cross-sentence hidden state, and $\mathbf{h}_{i} \in \vmathbb{R}^{d}$ denotes the representation of the report $d_i$. Max denotes the max pooling, [;] denotes concatenation, and $\theta$ denotes encoder parameters. 

\subsection{Report Selection}
Since this task is formulated on massive raw reports, our model aims to automatically narrow down the evidence extraction by ranking them  
and capturing the top ones for further analysis. 
Taking the claim in Fig. 1 as an example, there are $n$ retrieved reports about "microwaving fabric masks" and the significant reports $R_1$ and $R_n$ containing oracles (i.e., underlined sentences) are selected for veracity prediction and explanation generation. 

To distill the check-worthy reports from massive reports $\mathcal{D}$ that are helpful for veracity prediction, 
we firstly develop a coarse-grained document selector by treating the claim as a query to find $K$ most significant results. Then, global attention is utilized to obtain the significance score for each report $d_i$:
\begin{flalign}
\alpha_{c \rightarrow \mathcal{D}} &= \text{softmax} (\mathbf{H}_{\mathcal{D}} W_\alpha \mathbf{h}_c) \label{eq:doc_softmax}
\end{flalign}
where 
$\mathbf{H}_{\mathcal{D}} = [\mathbf{h}_{1};\mathbf{h}_{2};\cdots;\mathbf{h}_{{\mathcal{|D|}}}]$ compacts all hidden vectors of reports and $W_\alpha \in \vmathbb{R}^{d \times d}$ is a trainable parameter. 
We use $\alpha_{c \rightarrow \mathcal{D}} $ to rank all reports and select the top-$K$ results as the check-worthy reports  
(i.e., $ \hat{y}_{i}^{d} = \alpha_i (\alpha_i \geq \alpha_K)$ and otherwise $ \hat{y}_{i}^{d} = 0 (\alpha_i < \alpha_K)$). Note that the $t$-th sentence representation in the $k$-th selected report $d^{'}_k$ are denoted as $\mathbf{h}^{'}_{k,t} \in  \{\mathbf{h}^{'}_{k,1}, \mathbf{h}^{'}_{i,2}, ..., \mathbf{h}^{'}_{k,|d^{'}_k|} \}$, and its document representation is denoted as $\mathbf{h}^{'}_{k}$, which are used for explainable sentence extraction.

\subsection{Explainable Sentence Extraction}
On top of selected reports, we treat explanation generation as a multi-document extractive summarization, where each report is visited sequentially for explainable sentences. 
Such reports are regarded as the wisdom of crowds when detecting a dubious claim.  
We assume that explainable sentences for verdicts should be claim-relevant, informative, salient, and non-redundant. Specifically, there may exist redundancy between reports because a report is generally self-contained and multiple raw reports are more likely to contain semantically irrelevant and redundant sentences~\cite{ma2019sentence}. 

In this paper, 
we develop a fine-grained sentence selector to  extract explainable sentences from these check-worthy reports considering the following four  features: 1) \textit{claim relevance} measures the topic coverage of each sentence regarding the claim; 2) \textit{richness} measures the content informativeness of each sentence containing evidence; 3) \textit{salience} measures the significance of each sentence regarding the entire report; 4) \textit{non-redundancy} measures the novelty of each sentence regarding previous selected explainable sentences. 
Therefore, we define a layer to predict the probability of each sentence that should be selected via integrating the four features as follows: 
\begin{flalign}
\ \text{P}(y_{k,t}^s & = 1 | \mathbf{h}_c, \mathbf{h}^{'}_{k,t}, \mathbf{h}^{'}_{k}, \mathbf{h}_{d}) \nonumber \\
&= \sigma (\underbrace{\mathbf{h}^{'}_{k,t} W_c \mathbf{h}_c}_{(\textit{claim relevance})} +  \underbrace{\mathbf{h}^{'}_{k,t} W_s}_{(\textit{richness})}  \nonumber \\ 
 & + \underbrace{\mathbf{h}^{'}_{k,t} W_r \mathbf{h}^{'}_k }_{(\textit{salience})} - \underbrace{\mathbf{h}^{'}_{k,t} W_d \mathbf{h}_{d} }_{ (\textit{non-redundancy})}  )
\label{eq:explain_prob}
\end{flalign}
where $y_{k,t}^s$ is a binary variable indicating whether the $t$-th sentence in the selected report $d^{'}_k$ should be selected as part of explanations $\hat{E}$, 
and $W_*$ are trainable parameters. 
$\mathbf{h}_{d}$ is the redundancy vectors initialized with all zeros and updated by selected sentences in previously visited reports as follows: 
\begin{flalign}  
\mathbf{h}_{d} = \tanh ( \sum_t \mathbf{h}^{'}_{k-1,t} \cdot \text{P}(y_{k,t}^s = 1))  
\end{flalign} 

Considering the number of report sentences, 
our model learns to select the explainable sentences with probabilities above a soft threshold $\varepsilon_k=1 / |d_k^{'}|$, i.e., $\text{P}(y_{k,t}^s = 1) > \varepsilon_k$, where $\text{P}(y_{k,t}^s = 1)$ is obtained by Eq.  (\ref{eq:explain_prob}). 
Note that $\mathbf{h}^{''}_{k,t}$ is used to denote the sentence representation output from the explainable sentence selector.   

\subsection{Veracity Prediction}
To enhance final veracity prediction, we further employ the extracted explanation as additional \textit{evidence} besides the claims and all reports. 
Specifically, we aggregate the recognitions from such evidence and reports for a target claim, respectively, and then obtain the final representation by concatenating the claim representation, report representation, and explanation representation as follows: 
\begin{flalign}
\mathbf{h}_{D} &= \text{Max} ( [\mathbf{h}_{1}; \mathbf{h}_{2}; \cdots; \mathbf{h}_{|\mathcal{D}|}]) \\ 
\mathbf{h}_{E} &= \text{Max} ([\mathbf{h}^{''}_{1}; \mathbf{h}^{''}_{2}; ...; \mathbf{h}^{''}_{K}]) \\  
\mathbf{h}^{\dagger} &=  [\mathbf{h}_{c}; \mathbf{h}_{D}; \mathbf{h}_{E}]
\end{flalign}
where $\mathbf{h}_{D}$ denotes the integrated representation of all report sentences, $\mathbf{h}_{E}$ denotes the integrated representation of all explainable sentences. $\mathbf{h}^{\dagger}$ denotes the final representation for veracity prediction. 
$K$ denotes a hyperparameter controlling the maximum number of selected reports. 
Similar to Eq.  (\ref{eq:lstm}), 
$\mathbf{h}^{''}_k = \text{Max} ([\mathbf{h}^{''}_{k,1}; \mathbf{h}^{''}_{k,2}; \cdots; \mathbf{h}^{''}_{k,|d^{'}_k|}])$ is the $k$-th report representation in the extracted explanations. 

Finally, $\mathbf{h}^{\dagger}$ is fed 
into a multi-layer perceptron (MLP) layer to predict the veracity label as follows: 
\begin{flalign}
\hat{y} &= \text{softmax} (\text{MLP}(\mathbf{h}^{\dagger})) \label{eq:veracity} 
\end{flalign}


\subsection{Model Training}
It is inefficient to train report selection, explainable sentence extraction, and veracity prediction independently, considering their implicit correlations and the pipeline for explainable fake news detection in the real world \cite{kotonya2020survey}. 
Thus, we jointly optimize these three sub-tasks in an end-to-end model. For model training, we minimize the overall loss $\mathcal{L}_{all}$ as follows: 
%
\begin{flalign}
\mathcal{L}_D &= - \sum_i y_i^{d} \text{log}(\hat{y}_i^{d}) \label{eq:doc_loss} \\
\mathcal{L}_S &= - \sum_k \sum_t y_{k,t}^s \text{log}(\hat{y}_{k,t}^s) \label{eq:sent_loss} \\
\mathcal{L}_C &= - y \text{log}(\hat{y}) \label{eq:ver_loss} \\
\mathcal{L}_{all} &= \beta_D \mathcal{L}_D + \beta_S \mathcal{L}_S + \beta_C \mathcal{L}_C 
\label{eq:joint_loss}
\end{flalign}
where $\mathcal{L}_D$, $\mathcal{L}_S$, and $\mathcal{L}_C$ denote the cross-entropy loss for check-worthy report selection, explanation generation and veracity prediction tasks, respectively. 
$y_i^{d}$ and $\hat{y}_i^{d}$ denote the gold and predicted label of reports, respectively. 
$y_{k,t}^s,$ and $\hat{y}_{k,t}^s$ denote the ground truth and the predicted probability of the sentence for explanation, respectively. 
$y$ and $\hat{y}$ denote the ground truth and predicted veracity probability of the claim, respectively.  
$\beta$ denotes the trade-off parameter, 
controlling the task importance in our work. 
We can automatically assign $\beta_D, \beta_S$, and $\beta_C$ with proper values using the adaptive strategy, rather than the grid search (see Appendix \ref{app:MAW}). 



\section{Experiments}
\subsection{Datasets and Settings} 
To the best of our knowledge, there is no public dataset on raw reports available for this task. 
Thus, we collect two explainable datasets, i.e., RAWFC and LIAR-RAW, referring to two different fact-checking sites (i.e., Snopes\footnote{www.snopes.com} and Politifact\footnote{www.politifact.com}) for gold labels, respectively.  
For RAWFC, we constructed it from scratch by collecting the claims from Snopes and \textit{relevant raw reports} by retrieving claim keywords. 
For LIAR-RAW, we extended the public dataset LIAR-PLUS ~\cite{alhindi2018liarplus} with \textit{relevant raw reports}, containing \textit{fine-grained} claims from  Politifact. 
We process and separate these datasets into train/valid/test sets by 8:1:1 following the same setting in \cite{atanasova2020generating}. More details are illustrated in Appendix~\ref{app:data}. 
\renewcommand{\arraystretch}{0.6} 
\begin{table}[t!]
\normalsize 
	\centering
	\setlength{\tabcolsep}{4.0mm}
	\resizebox{1.0\linewidth}{!}{
	\begin{tabular}{l r  r }
		\toprule
		Dataset       &RAWFC  &LIAR-RAW      \\
		\midrule
		Claim                       & 2,012     & 12,590    \\ 
	     \quad\# pants-fire         & -         & 1,013      \\%
	     \quad\# false               & 646      & 2,466   \\%
	     \quad\# barely-true        & -         & 2,057     \\
	     \quad\# half-true $\dag$   & 671       & 2,594     \\
	     \quad\# mostly-true        & -         & 2,439     \\
	     \quad\# true               & 695       & 2,021     \\
	   Veracity Label            & 3         & 6    \\
	   \midrule
		Explain sentence     \\ 
			\quad \# min    & 1      & 1 \\
		    \quad \# max    & 110       & 209     \\
		    \quad \# avg    & 18.4      & 4.1     \\
	   \midrule 
		Report per claim     \\ %
		    \quad \# min  & 1        & 1  \\ 
		    \quad \# max  & 30       & 30  \\
		    \quad \# avg  & 21.0    & 12.3     \\
		Sentence per report    \\ 
		    \quad \# min    & 1     & 1       \\
		    \quad \# max    & 155   & 59   \\
		    \quad \# avg    & 7.4   & 5.5   \\
		\bottomrule
	\end{tabular}
	}
	\caption{Statistics of datasets. \# half-true $\dag$ is also denoted as \# half in RAWFC. The number of oracles in datasets isn't pre-defined. } %
	\label{tab:Statistics} \vspace{-0.5cm}
\end{table}


\renewcommand{\arraystretch}{0.64} 
\begin{table*}[t!] 
\normalsize 
	\centering 

	\resizebox{0.9\linewidth}{!}{
		\begin{tabular}{l c c c  c c c} 
			\toprule
			
			\multirow{2}{*}{Model}& 
			\multicolumn{3}{c}{RAWFC}&\multicolumn{3}{c}{LIAR-RAW}\\
			\cmidrule(lr){2-4} \cmidrule(lr){5-7} &P(\%)&R(\%)&macF1(\%) &P(\%)&R(\%)&macF1(\%)\\
			\midrule 
			SVM \cite{pedregosa2011scikit}        	&32.33 &32.51 &31.71                &15.78  &15.92  &15.34  \\
			CNN \cite{wang2017liar}                 &38.80 &38.50 &38.59 	            &22.58 &22.39 &21.36 \\
			RNN \cite{rashkin2017lstm}   	        &41.35 &42.09 &40.39	            &24.36 &21.20 &20.79 \\%
			DeClarE \cite{popat2018declare}         &43.39 &43.52 &42.18 	            &22.86 &20.55 &18.43  \\
			dEFEND \cite{shu2019defend}      		&44.93 &43.26 &44.07                &23.09 &18.56 &17.51  \\%
			SentHAN \cite{ma2019sentence}      		&45.66 &45.54 &44.25 	            &22.64 &19.96 &18.46  \\%
			SBERT-FC \cite{kotonya2020health}       &51.06 &45.92 &45.51	            &24.09  &22.07 &22.19  \\ 	
			GenFE \cite{atanasova2020generating}    &44.29 & 44.74 &44.43	            &28.01  &26.16 &26.49  \\ 	
			GenFE-MT \cite{atanasova2020generating} &45.64 & 45.27 &45.08	            &18.55  &19.90 &15.15  \\	%
            CofCED                  &\textbf{52.99}  &\textbf{50.99}  &\textbf{51.07}		 		&\textbf{29.48}  &\textbf{29.55}  &\textbf{28.93}		\\ 
			\bottomrule 
		\end{tabular} 
    	}
	\caption{Experimental results of veracity prediction merely using raw reports ($p<0.05$ under t-test).  
	} %
	\label{tab:veracity-acc} \vspace{-0.4cm}
\end{table*} 

For experimental setup, we initialized word embeddings with the base uncased DistilBERT~\cite{sanh2019distilbert} and $d=768$ dimensions. The hidden size of LSTM is set to 384. We use Adam optimizer \cite{kingma2014adam} with a learning rate of 1e-5 and the mini-batch size is set to 1 to minimize joint cross-entropy loss. The maximum number $K$ of selected reports for each claim is empirically set to 12 and 18 for RAWFC and LIAR-RAW, respectively. We use a soft threshold $\varepsilon_i=1/|d_i^{'}|$ for selection while empirically setting the maximum number of oracle sentences to 30 and 55 for RAWFC and LIAR-RAW, respectively. 
We set the dropout rate to 0.4 before final prediction and the maximum number of training epochs to 8. 
For evaluation, we employ macro-averaged precision (P), recall (R), and F1 score (macF1) for veracity prediction, and use ROUGE-$N$ F1 score ($N \in \{1, 2, L\}$) and the human evaluation to evaluate the quality of explanations. \textit{Note that fact-checked reports are not required during inference in our model}.

\subsection{Veracity Prediction Performance}
Table~\ref{tab:veracity-acc} compares veracity prediction results with the following strong baselines: 1) \textbf{SVM} \cite{pedregosa2011scikit}: This uses bag-of-words features to train SVM-based model for fake news detection; 2) \textbf{CNN} \cite{wang2017liar}: This incorporates available metadata features to enhance representation learning; 3) \textbf{RNN} \cite{rashkin2017lstm}: This learns representation from word sequences without external resources; 4) \textbf{DeClarE} \cite{popat2018declare}: This combines word embeddings from the claim, report, and source to access the credibility of the claim; 
5) \textbf{dEFEND}~\cite{shu2019defend}: This utilizes GRU-based model for veracity prediction with explanations; 
6) \textbf{SentHAN} \cite{ma2019sentence}: This represents each sentence based on sentence-level coherence and semantic conflicts with the claim; 7) \textbf{SBERT-FC} \cite{kotonya2020health}: This uses SentenceBERT (SBERT) for encoding and detects fake news based on the top-$K$ ranked sentences; 8) \textbf{GenFE/GenFE-MT} \cite{atanasova2020generating}: This detects fake news independently or jointly with explanations in the multi-task set-up.   

Table~\ref{tab:veracity-acc} demonstrates the detection performance of our proposed CofCED compared with existing strong baselines in terms of precision, recall and macro F1 (macF1). From this table, we can observe that CNN and RNN outperform SVM on both datasets, indicating that deep learning methods can better capture semantic and syntactic features from raw reports. By attentively aggregating multiple features from the claim, reports, and source to estimate the veracity, dEFEND, DeClarE and SentHAN achieve better performance on RAWFC but slightly worse results on LIAR-RAW, because fine-grained labels contained in LIAR-RAW make it more challenging.

SBERT-FC and GenFE outperform SentHAN and dEFEND on both datasets, demonstrating the superiority of pre-trained models. GenFE-MT performs better than GenFE on RAWFC, but much worse than other baselines on LIAR-RAW, implying the challenge of fine-grained fake news detection with explanation generation in the multi-task setting. 
Generally, CofCED consistently achieves much better performance on RAWFC and LIAR-RAW, demonstrating the superiority of CofCED in combining report selection, explainable sentence extraction and veracity prediction for fake news detection directly on raw reports, 
alleviating the dependency on fact-checked reports. 

\renewcommand{\arraystretch}{0.66} 
\begin{table*}[t!] 
\normalsize 
	\centering 

	\resizebox{0.76\linewidth}{!}{
		\begin{tabular}{l c c c  c c c} 
			\toprule
			
			\multirow{2}{*}{Model}& 
			\multicolumn{3}{c}{RAWFC}&\multicolumn{3}{c}{LIAR-RAW}\\
			\cmidrule(lr){2-4} \cmidrule(lr){5-7} &P(\%)&R(\%)&macF1(\%) &P(\%)&R(\%)&macF1(\%)\\
            \midrule
            CofCED w/o RS\&SE       &45.01  &45.02  &44.98 		 	&25.69  &24.55  &24.80		\\ 
            CofCED w/o SE           &52.27  &46.36  &43.80             &27.59  &23.81  &23.74 \\ 
            CofCED w/o RS           &49.26  &46.92  &46.37		 		&27.08  &25.32  &25.52	\\ 
            CofCED w/o non-redundancy  &48.80  &46.98  &47.48		 	&26.54  &27.36  &26.65	\\ 
            CofCED w/o salience        &43.96  &49.24  &46.44		 	&26.36  &24.88  &25.23	\\ 
            CofCED w/o richness        &48.08  &47.50  &47.12		 	&27.06  &25.82  &26.05	\\ 
            CofCED w/o claim relevance       &45.66  &45.25  &45.28		 	&26.42  &24.01  &24.88	\\ 
            CofCED                  &\textbf{52.99}  &\textbf{50.99}  &\textbf{51.07}		 		&\textbf{29.48}  &\textbf{29.55}  &\textbf{28.93}		\\ 
			\bottomrule 
		\end{tabular} 
    	}
	\caption{Ablation study results of our veracity prediction on test sets; w/o denotes `without'. 
	} 
	\label{tab:ablation} \vspace{-0.2cm}
\end{table*} 

\renewcommand{\arraystretch}{0.6} 
\begin{table*}[t!] 
\normalsize 
	\centering 

	\resizebox{0.86\linewidth}{!}{
		\begin{tabular}{l r r r  r r r} 
			\toprule
			
			\multirow{2}{*}{Model} 
			& \multicolumn{3}{c}{RAWFC} &\multicolumn{3}{c}{LIAR-RAW}\\			\cmidrule(lr){2-4} \cmidrule(lr){5-7}  &ROU-1 &ROU-2 &ROU-L  &ROU-1 &ROU-2 &ROU-L\\
			\midrule
			LEAD-N                              &19.52 & 4.54 & 17.26      &9.84 & 0.40 & 7.20     	\\ 
			Oracle                              &37.62 &13.22 &34.67 	   &25.50 &9.28 &22.61 \\%
			\midrule
			EXTABS \cite{kotonya2020health}   &- &- &-	               &18.85 &3.61 &12.90 	\\ 
			dEFEND \cite{shu2019defend}  &19.95 &5.08 &17.21   	&17.03 &3.26 &11.42\\%
			GenFE-MT \cite{atanasova2020generating}  &18.23 &7.12 &17.32  	    &\textbf{23.08} &3.67 &12.10   \\ 
            \midrule
            CofCED w/o non-redundancy   &27.32 & 9.06 & 23.19		 	&17.96  &3.54  &12.43	\\ 
            CofCED w/o salience     &26.67 & 7.44 & 21.02		 		&17.27  &3.41  &11.69	\\ 
            CofCED w/o richness     &25.75 & 8.66 & 21.87		 		&17.23  &3.44  &12.10	\\ 
            CofCED w/o claim relevance    &25.56 & 8.07 & 20.73		&17.08  &3.31  &11.25	\\ 
            CofCED w/o RS           &26.64 & 8.96 & 22.69        &17.51  &\textbf{3.72} &\textbf{13.20}  \\%
            CofCED                  &\textbf{27.62} &\textbf{9.32} &\textbf{23.57}       &17.14  &3.49  &12.96   \\
			\bottomrule 
		\end{tabular} 
	}
	\caption{ROUGE results of the generated explanation. ROU-$N$ ($N \in \{1, 2, L\}$) denotes the ROUGE-$N$ F1 score that evaluates the token overlap between the explanation and human justifications. RAWFC is not suitable for EXTABS because its gold justification is too long to train an abstractive-summarization model. 
	} %
	\label{tab:exp-rouge} \vspace{-0.4cm}
\end{table*} 
\subsection{Ablation Study}
To evaluate the impact of each component, we conduct ablation experiments for CofCED by removing the following key components: 1) \textbf{RS} denotes report selection; 2) \textbf{SE} denotes sentence selection; 3) \textbf{RS\&SE} denotes RS and SE; 4) Four semantic features: \textbf{claim relevance}, \textbf{richness}, \textbf{salience}, and \textbf{non-redundancy}, for sentence selection.  

As shown in Table~\ref{tab:ablation}, CofCED significantly outperforms CofCED w/o $*$ ($*$ indicates a component) on both datasets, demonstrating all components contribute to the effectiveness of CofCED in detecting fake news. Specifically, 
CofCED's performance significantly decreases without RS\&SE because there is noise in raw reports, affecting the veracity prediction. 
CofCED w/o SE performs much worse than the others because irrelevant or redundant information contained in such reports may weaken the effect of evidence for detection; 
CofCED w/o RS also achieves worse performance than CofCED because noisy reports may affect sentence selection and model training.
Furthermore, the performance of CofCED w/o claim relevance significantly decreases, highlighting the importance of selecting claim-relevant evidence for final prediction. CofCED outperforms CofCED without these four features for sentence selection, respectively, demonstrating they contribute to extracting explainable sentences for fake news detection from different perspectives.   

\subsection{Explanation Evaluation}
Table~\ref{tab:exp-rouge} reports the ROUGE results of the extracted explanations regarding word overlapping. The ROUGE F1 score is employed to evaluate their qualities 
comparing with the following strong baselines: 1) \textbf{LEAD-N} \cite{nallapati2017summarunner}: This uses the first N sentences as explanation and $N = 5$; 
2) \textbf{Oracle}~\cite{atanasova2020generating}: This typically presents the best greedy approximation of the gold explanation with sentences extracted from reports; 3) \textbf{EXTABS} \cite{kotonya2020health}: This uses extractive-abstractive summarization model pre-trained on extra news articles and summaries dataset before fine-tuning~\cite{liu2019text};  4) \textbf{dEFEND}: This uses internal attention weights for explanations; 5)  \textbf{GenFE-MT}: This  incorporates explanation generation using pre-trained models. 

\renewcommand{\arraystretch}{0.8} 
\begin{table*}[ht!] 
\small 
	\centering 
		\begin{tabular}{p{0.76\linewidth} p{0.01\linewidth} p{0.01\linewidth} p{0.01\linewidth} p{0.01\linewidth} p{0.01\linewidth} p{0.01\linewidth}}
			\toprule 
			\textbf{Claim:} \textit{Dr. Tasuku Honjo said that COVID-19 was ``man-made" at a lab in Wuhan, China.} & \multirow{2}{*}{\rotatebox{90}{Relevance}} & \multirow{2}{*}{\rotatebox{90}{Richness}} & \multirow{2}{*}{\rotatebox{90}{Salience}} & \multirow{2}{*}{\rotatebox{90}{Non-redant}} & \multirow{2}{*}{\rotatebox{90}{\textbf{Overall}}} & \multirow{2}{*}{\rotatebox{90}{}} \\ 
			$[$\textbf{Prediction:} False$]$ \textbf{Explanation:} 
			Honjo did not work at the Wuhan Institute of Virology, he did not say that COVID-19 was “invented” or “man-made,” and the Twitter account posting similar claims does not belong to the Nobel Prize winner. In addition, this rumor is all based on the unfounded notion that COVID-19 was created as a bioweapon. (...) \\
			\midrule
            $[1]$ TOKYO, May 6 (Xinhua) -- Japanese Nobel laureate Tasuku Honjo have refuted claim that China manufacture the novel coronavirus, say those rumor be ``dangerously distract.''  & 0.9 & 0.6  &0.8 & 0.9    & \textbf{0.9} &$\surd$ \\
            $[2]$ Actually, the professor don't have a Twitter account.     &0.7 &0.5  &0.6 & 0.9    & \textbf{0.6} &$\surd$\\%
            $[3]$ The 2018 Nobel laureate encourage Japanese authority to adopt a more proactive approach.      &0.3 &0.5  &0.4 & 0.8    & \textbf{0.3} &\texttimes \\%
            $[4]$ China will have a big role to play. ...         &0.2 &0.2  &0.1 & 0.7    & \textbf{0.2} &\texttimes \\%
			\bottomrule 
		\end{tabular} 
	\caption{
	Our visualization of explanation extraction from raw reports. Each row is a sentence in raw reports. The score in the columns are normalized from each of the abstract features in Eq.  (\ref{eq:explain_prob}), and the last column is the final probability explaining to detection results.   
	} 
	\label{tab:case-analysis} \vspace{-0.6cm}
\end{table*} 

Overall, CofCED achieves the state-of-the-art performance on RAWFC and comparable ROUGE scores with GenFE-MT on LIAR-RAW, suggesting that our CofCED can effectively distill explainable sentences that contributes to the final veracity prediction, as shown in Table \ref{tab:veracity-acc}. 
Specifically, the ROUGE results of LEAD-N and Oracle on RAWFC and LIAR-RAW indicate that generating explanations for fine-grained fake news detection is a more complex challenge. 
EXTABS obtains competitive results on LIAR-RAW due to additional news and summaries datasets for abstractive summarization but it cannot deal with long justifications. 
GenFE-MT performs much better than dEFEND on both datasets, indicating the advantage of pre-trained models in generating explanation from raw reports but failing to trade off both tasks regarding Table~\ref{tab:veracity-acc}.  
For ablation results, we observe that some ablations of CofCED achieve slightly better ROUGE scores but much worse veracity predictions on LIAR-RAW, indicating these four features can effectively select explainable sentences to enhance fake news detection. 
Besides, CofCED performs better on RAWFC while only comparable on LIAR-RAW than CofCED w/o RS, 
implying that generating explanations for fine-grained veracity labels is much more challenging regarding word overlapping.  
We further conduct human evaluations as shown in Appendix \ref{app:human}. 
In summary, our CofCED can effectively generate accurate explanations from raw reports and all components contribute to focusing on veracity prediction.  

\subsection{Case Study} 
For in-depth analysis, 
we further explore the process of CofCED in selecting explainable sentences. We normalized scores for each abstract feature, obtaining its overall probability for explaining detection results. As shown in Table~\ref{tab:case-analysis}, given a false claim about COVID-19,  
the top two sentences with higher overall scores refute the claim from different perspectives and the last two sentences with 0.3 and 0.2 overall scores contribute less to the veracity prediction. The separated terms, i.e., claim relevance, richness, salience, and non-redundancy, in  Eq.  (\ref{eq:explain_prob}) are clearly visualized for seeking the major factor responsible for the classification of each sentence.  
In addition to being a state-of-the-art method for explainable fake news detection, CofCED has the additional superiority of being very explainable for sentence extraction. Thus, such visualization increases the transparency of the system and the credibility of generated explanations for verdicts. 
\begin{figure}[t!]
\centering
\subfloat[Veracity prediction]{
\begin{minipage}[t]{0.5\linewidth}
\centering
\includegraphics[width=1.0\linewidth]{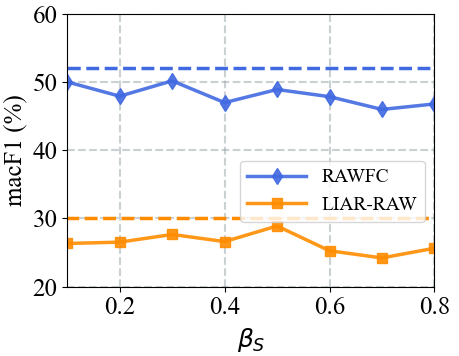}
\end{minipage}%
}%
\subfloat[Explanation generation]{
\begin{minipage}[t]{0.5\linewidth}
\centering
\includegraphics[width=1.0\linewidth]{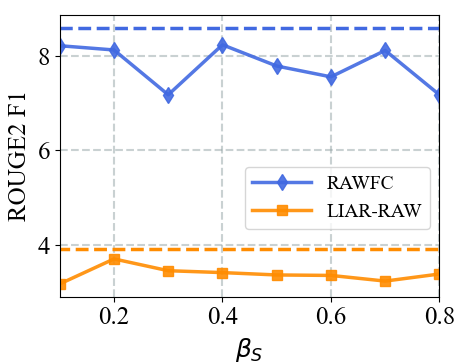}
\end{minipage}%
}
\centering
\caption{
Results of CofCED under different values of the trade-off parameter $\beta_S$ and 
$\beta_C = 1 - \beta_S$.  The colored dashed horizontal lines denote the performance of CofCED with our adaptive weighting. 
}
\label{fig:impact_tradeoff} 
\vspace{-0.5cm}
\end{figure} 
\begin{figure}[t!]
\centering
\subfloat[Veracity prediction]{
\begin{minipage}[t]{0.5\linewidth}
\centering
\includegraphics[width=1.0\linewidth]{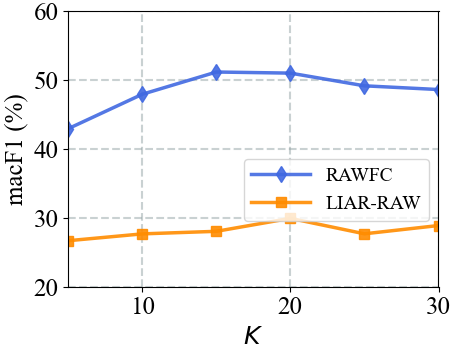}
\end{minipage}%
}%
\subfloat[Explanation generation]{
\begin{minipage}[t]{0.5\linewidth}
\centering
\includegraphics[width=1.0\linewidth]{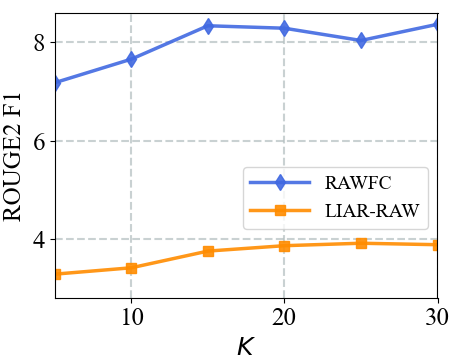}
\end{minipage}%
}
\centering
\caption{Results of CofCED under different values of the maximum number $K$ for report selection. } 
\label{fig:impact_topk} 
\vspace{-0.4cm}
\end{figure} 
\subsection{Parameter Sensitivity Study}
We further investigate the impact of the trade-off parameter $\beta$ in Eq.  (\ref{eq:joint_loss}) on CofCED using the grid search. For brevity, Fig.~\ref{fig:impact_tradeoff} only presents the results for a) veracity prediction and b) explanation generation on development sets  
when $\beta_S$ varies and $\beta_D=0.5$ is temporarily fixed. 
We also tried various $\beta_D \in [0.1, 0.8]$ and consistently achieved similar results.  
By varying the value of $\beta_S$ from 0.1 to 0.8, our model achieves better performances on one task but poorer results on the other. This is because these tasks show different importance and priority for the final performance over time. By contrast, our CofCED with our proposed multi-task adaptive weighting (MAW) (i.e., the colored dashed horizontal lines) consistently achieves better performance.  
Thus, these results demonstrate that CofCED with MAW can effectively find better weights for explanation generation and veracity prediction in multi-task learning, alleviating the labor for the grid search for trade-off parameters.

To examine the impact of the maximum number of selected reports on CofCED, we conduct experiments by varying $K$ while fixing other hyper-parameters on the development sets of RAWFC and LIAR-RAW. As shown in Fig.~\ref{fig:impact_topk}, we can see that too few raw reports generally cause performance reduction because the noise in the raw reports may impose the model training bias. Since too many raw reports will cause the out of memory problem, we empirically choose a proper value in this study, i.e., $K$ is set to 12 and 18 for RAWFC and LIAR-RAW, respectively. 
Note that $\varepsilon$ is a soft threshold that can be automatically assigned regarding the total number of report sentences.

\section{Conclusion}
We present a coarse-to-fine cascaded evidence-distillation (CofCED) neural network for explainable fake news detection that achieves the best detection performance and distills accurate veracity explanations directly from raw reports.  
Besides, CofCED has the additional advantage of 
being explainable in producing veracity explanations, explicitly considering the semantic features, e.g., claim relevance, richness, salience, and non-redundancy.  Experimental results on real-world datasets demonstrate 
the effectiveness of CofCED for explainable fake news detection utilizing the wisdom of crowds, effectively mitigating the dependency on fact-checked reports.   
%


\section*{Acknowledgments}
This work is partially supported by National Natural Science Foundation of China through grants No.61976102, No.U19A2065, and No.61902145. This work is partly supported by the International Cooperation Project (20220402009GH) and Science \& Technology Development Program (20210508060RQ), Jilin Province. 
This work is partly supported by HKBU One-off Tier 2 Start-up Grant (RCOFSGT2/20-21/SCI/004), Hong Kong RGC ECS (22200722).



\bibliography{anthology,coling22}
\bibliographystyle{acl_natbib}

\clearpage
\clearpage
\appendix    
\section*{Appendices}
\label{appx}
\setcounter{subsection}{0}
\setcounter{equation}{0}
\setcounter{table}{0}
\setcounter{figure}{0}
\renewcommand{\thesubsection}{A.\arabic{subsection}}
\renewcommand{\theequation}{A.\arabic{equation}}
\renewcommand{\thetable}{A.\arabic{table}}
\renewcommand{\thefigure}{A.\arabic{figure}}

\section{Dataset Details}
\label{app:data}
Existing benchmarks for explainable fake news detection collected official debunked reports written by journalists as evidence for fake news detection \cite{kotonya2020survey}, which is labor-intensive and relatively inefficient. However, debunked reports are not always available for breaking news and are mixed up with raw reports,  
which may contain more semantically irrelevant and redundant information. 
To the best of our knowledge, there is no available explainable dataset based on crowds of raw reports to detect fake news before official reports published. Thus, existing datasets are not suitable for most real-life scenarios, especially when 
the fact-checked reports are not always available. To address this issue, we collect two new datasets, i.e., RAWFC and LIAR-RAW, considering a more general situation of detecting and explaining fake news with relevant raw reports. 

Note that we construct RAWFC and LIAR-RAW with gold labels referring to Snopes\footnote{www.snopes.com} and  Politifact\footnote{www.politifact.com}, respectively. RAWFC is constructed from scratch as follows and LIAR-RAW are extended with raw reports based on LIAR-PLUS~\cite{alhindi2018liarplus}. 
Besides, we pre-processed LIAR-RAW similar to RAWFC. 
The detailed statistics of datasets are shown in Table \ref{tab:Statistics}. 

\subsection{Data Collection and Processing.} 
We crawled claims with their veracity labels and relevant fact-checked reports that can be regarded as gold explanations from Snopes. For each claim, we extracted the claim-related keywords as the search query and used Google API to retrieve the top $30$ relevant raw reports. 
To mitigate the dependency on fact-checked reports, 
we filtered out reports from fact-checking sites and 
removed the raw reports published after the publication time of the fact-checked report. 
We further removed the summary from the remaining articles and improved the quality of the dataset with data cleanings, e.g., removing reports containing less than 5 words or more than 3000 words. 
Finally, we standardized the original labels for 3-way classification: \{\textit{true}, \textit{false}, \textit{half}\}, i.e., 
\{true, correct attribute, mostly true\} $\rightarrow$ \textit{true}, 
\{ false, misattribute, mostly false \} $\rightarrow$ \textit{false}, 
\{mixture, unproven \} $\rightarrow$ \textit{half}. 
Each sentence is annotated as evidence or not according to their similarities with the gold explanation, where we greedily extract sentences that achieve the high cosine similarity and ROUGE F1 score, referred to as \textit{oracles}. 
\begin{figure} [t!]
	\centering
	\includegraphics[width=0.95 \linewidth]{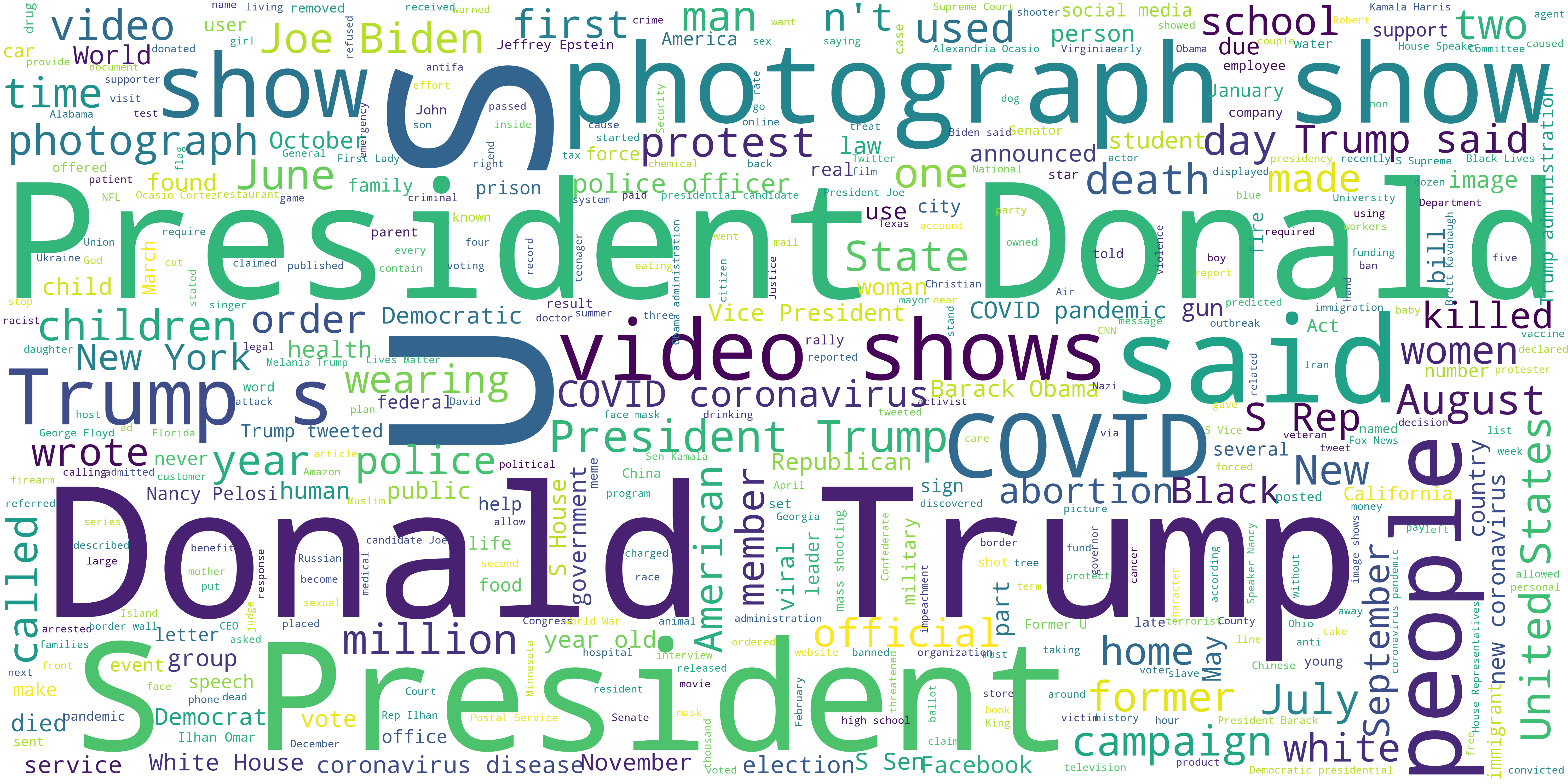} 
	\caption{The word cloud of our RAWFC. 
    }
    \label{fig:wordcloud1}\vspace{-0.2cm}
\end{figure}

\begin{figure} [t!]
	\centering
	\includegraphics[width=0.95 \linewidth]{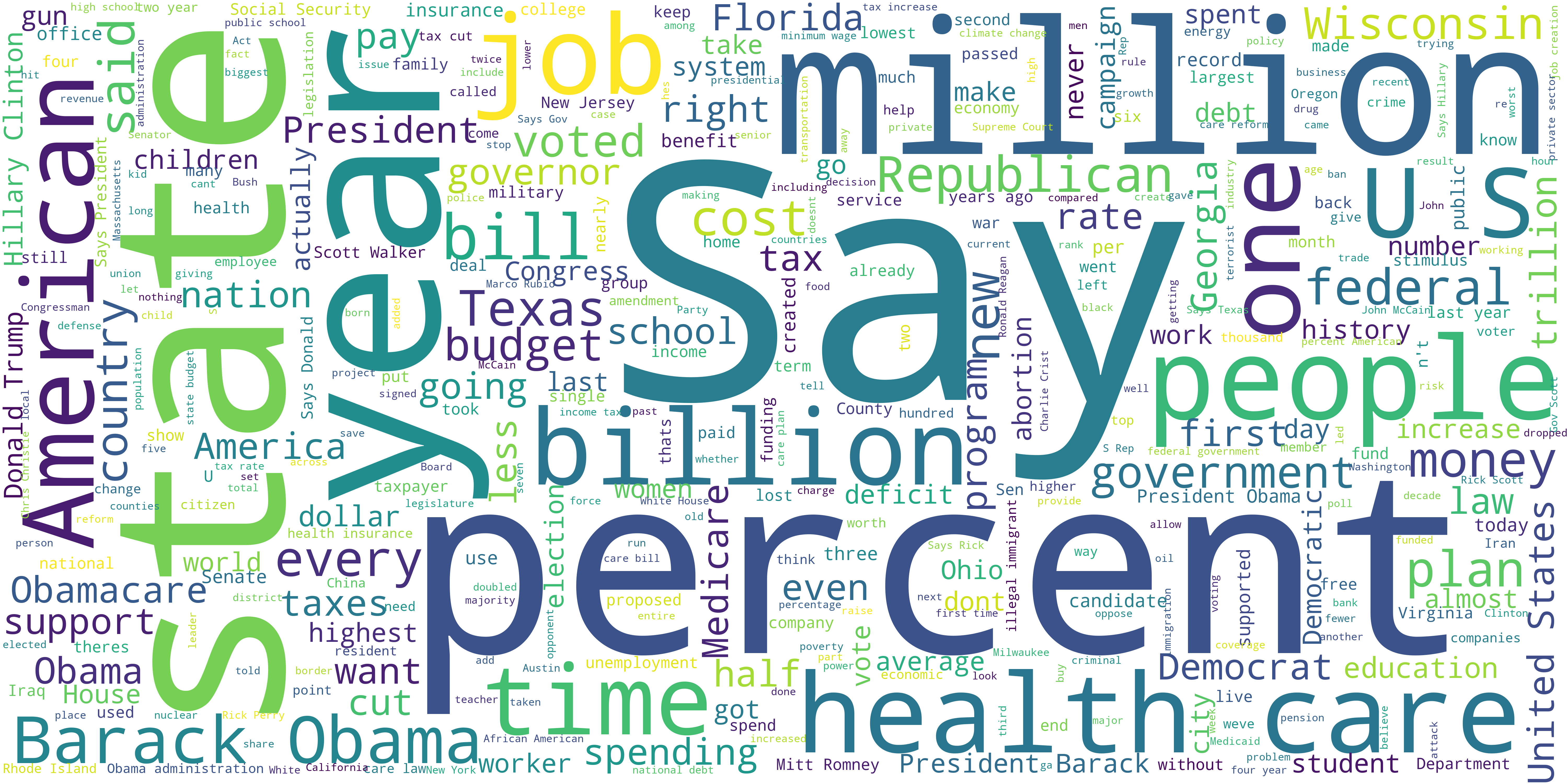} 
	\caption{The word cloud of our LIAR-RAW. 
    }
    \label{fig:wordcloud2}\vspace{-0.2cm}
\end{figure}

\subsection{Evidential Sentence Annotation.} 
To help produce explanations from external raw reports, each sentence in the article is annotated as evidence or not. 
Different from selecting evidential sentences based merely on the ROUGE score~\cite{lin-2004-rouge} with gold explanations \cite{atanasova2020generating}, we propose a more practical approach to annotate sentences according to both textual-level and semantic-level similarities.

For each candidate sentence, we adopt two metrics to assess whether it should be selected or not: 1) \textit{ROUGE} measures the textual-level similarity regarding the gold explanation in terms of the $n$-gram overlap; and 2) \textit{Cosine} measures the semantic similarity regarding the gold explanation. 
Formally, for a candidate sentence $s_{i,j} \in d_i = \{s_{i,j}\}_{j=1}^{|d_i|}$  
and its corresponding explanation sentences set $E=\{e_1,e_2,...,e_n\}$, we define the $n$-gram overlap function $f^{ROU}(s_{i,j},e_i)$  and semantic similarity $f^{COS}(s_{i,j}, e_i)$ as follows: 
\begin{align}
    f^{ROU}(s_{i,j},e_i)  &= \frac{|n\text{-grams}(s_{i,j})\ \cap \  n\text{-grams}(e_i)|}{ |n\text{-grams}(e_i)|} \\
    f^{COS}(s_{i,j}, e_i) &= \cos(h_{s_{i,j}}, h_{e_i}),
\end{align}
where $h_{s_{i,j}}$ and $h_{e_i}$ is the sentence representation encoded by SBERT~\cite{reimers2019SBERT}. 
We calculate the textual similarity in terms of \textit{ROUGE-1}, \textit{ROUGE-2}, and \textit{ROUGE-L} F1 scores, respectively; we also calculate the semantic similarity in terms of \textit{Cosine}. For sentence labeling, 
we empirically set the thresholds of \textit{ROUGE-1}, \textit{ROUGE-2}, and \textit{ROUGE-L} F1 scores to 0.1, 0.0, and 0.1, respectively, and the threshold of \textit{Cosine} to 0.6. Finally, we accepted the sentences that exceed all given thresholds as gold explanation sentences, i.e., \textit{oracle}.  The label statistics of claims in RAWFC and LIAR-RAW are displayed in Table \ref{tab:Statistics} and Table \ref{stat:LIAR-RAW}, respectively. Moreover, we also visualized their word clouds, as shown in Fig.  \ref{fig:wordcloud1} and Fig. \ref{fig:wordcloud2}, respectively. 


\begin{table}[t!]
    \centering
    \begin{tabular}{crrr}
        \toprule
        Standardized Label & Train & Valid & Test\\
        \midrule
        true  & 561 & 67 & 67 \\
        false & 514 & 66 & 66  \\
        half  & 537 & 67 & 67\\
        \bottomrule
    \end{tabular}
     \caption{Label statistics of claims in RAWFC.}
     \label{stat:RAWFC}
\end{table}
\begin{table}[t!]
    \centering
    \begin{tabular}{crrr}
        \toprule
        Fine-grained Label & Train & Valid & Test\\
        \midrule
        pants-fire   & 812   & 115 &86  \\        
        false        & 1,958 & 259 &249 \\
        barely-true  & 1,611 & 236 &210 \\
        half-true    & 2,087 & 244 &263 \\  
        mostly-true  & 1,950 & 251 &238 \\
        true         & 1,647 & 169 &205 \\
        \bottomrule
    \end{tabular}
     \caption{Label statistics of claims in  LIAR-RAW.}
     \label{stat:LIAR-RAW}
     \vspace{-0.2cm}
\end{table}
\section{Multi-task Adaptive Weighting}
\setcounter{subsection}{0}
\setcounter{equation}{0}
\setcounter{table}{0}
\setcounter{figure}{0}
\renewcommand{\thesubsection}{B.\arabic{subsection}}
\renewcommand{\theequation}{B.\arabic{equation}}
\renewcommand{\thetable}{B.\arabic{table}}
\renewcommand{\thefigure}{B.\arabic{figure}}

\label{app:MAW}
Inspired by prior work \cite{chen2018gradnorm,liu2019end}, 
we further propose a simple yet effective remedy, namely Multi-task Adaptive Weighting (MAW), to automatically keep a dynamic balance among tasks for different benchmark datasets. We define the weighting function $\beta_k(t)$ as follows:  
\begin{flalign}
\mathcal{\beta}_{k}(t) &=  \frac{N_k \exp[f_k(t) g(t)]}{\sum_i \exp[f_i(t) g(t)]} \label{eq:lumbda}\\
f_k(t) &= \frac{\mathcal{L}_k(t-1)}{\mathcal{L}_k(t-2)}, g(t) = \frac{\log(t-2)}{T} \label{eq:lossrate}
\end{flalign}
where $\beta_k = \beta_k(t), k \in \{\textit{D}, \textit{S}, \textit{C}\}$ and $f_k(t)$ represents the loss rate for task where $t$ is an iteration step; 
$g(t)$ is a global function that can generate a growth value, contributing to an optimal balance between tasks, since a large $T$ can result in a more even distribution between different tasks. $T=8$ denotes an initial temperature to control the softness of task weighting similar to \cite{caruana1997multitask}. 
$N_k=3$ indicates the total number of sub-tasks. 
We simply initialize $\beta_k=0.5$ and update the average loss over each iteration. 


\section{CofCED Algorithm}
\setcounter{subsection}{0}
\setcounter{equation}{0}
\setcounter{table}{0}
\setcounter{figure}{0}
\renewcommand{\thesubsection}{C.\arabic{subsection}}
\renewcommand{\theequation}{C.\arabic{equation}}
\renewcommand{\thetable}{C.\arabic{table}}
\renewcommand{\thefigure}{C.\arabic{figure}}

Algorithm \ref{algo} shows our training procedure. 
\renewcommand{\algorithmicrequire}{ \textbf{Input:}} 
\renewcommand{\algorithmicensure}{ \textbf{Output:}} 


\begin{algorithm}[t!]
\caption{CofCED}
\label{algo}
\LinesNumbered 
\KwIn{A set of training instances \{($c, \mathcal{D}$)\}; Maximum selection number $K$; Thresholds $\varepsilon$.}
\KwOut{Veracity label $\hat{y}$; Check-worthy report labels $\hat{Y}^d$; Explainable sentence labels $\hat{Y}^s$; Generated Explanation $\hat{E}$} 
        Initialize $\beta_D= \beta_S= \beta_C=0.5$, if $t \leq 2$; \\
        \For{each instance $(c, \{
        \{s_{i,j}\}_{j=1}^{|d_i| }\}_{i=1}^{|\mathcal{D}|} 
        )$ }
        {
        \algorithmiccomment{\textit{Hierarchical Encoding}}\\
        $\mathbf{h}_{c}, \mathbf{h}_{i,j} \gets$ DistilBERT; \\ 
        $\mathbf{h}_{i} \gets$ Eq. (\ref{eq:lstm});\\
        \algorithmiccomment{\textit{Task 1: Report Selection}}\\
        $\hat{y}_{i}^d, \{d^{'}_{k}\}_{k=1}^K \gets$ $K$; Eq.  (\ref{eq:doc_softmax}); \\
        $\mathbf{h}_{D} = \text{Max} ([\mathbf{h}_1; \mathbf{h}_2; ...; \mathbf{h}_{|\mathcal{D}|}]) $ \\
        
        \algorithmiccomment{\textit{Task 2: Explainable Sentence Extraction}}\\
        \For{ each report $d_k$ in $\{d^{'}_k\}_{k=1}^K$ }
            {
            $\hat{y}_{k,t}^s \gets $ Eq. (\ref{eq:explain_prob}); \\ 
            $\{s_{k, t}\}_{t=1}^{|d^{'}_k|}, \{ \mathbf{h}^{''}_{k, t} \}_{t=1}^{|d^{'}_k|} \gets \hat{y}_{k,t}^s > \varepsilon_k$ \\
            }
        Explanations: 
        $\hat{E}=\{ \{s_{k,t}\}_{t=1}^{|d^{'}_k|}  \}_{k=1}^{K}$, \\
        $\mathbf{h}^{''}_k = \text{Max} ([\mathbf{h}^{''}_{k,1}; \mathbf{h}^{''}_{k,2}; \cdots; \mathbf{h}^{''}_{k,|d^{'}_k|}])$; \\ 
        $\mathbf{h}_{E} = \text{Max} ( [\mathbf{h}^{''}_{1}; \mathbf{h}^{''}_{2}; ...; \mathbf{h}^{''}_K]) $; \\ 
        
        \algorithmiccomment{\textit{Task 3: Veracity Prediction}}\\
        $\mathbf{h}^{\dagger} =  [\mathbf{h}_{c}; \mathbf{h}_{D}; \mathbf{h}_{E}]$\\
        Verdicts: $\hat{y} \gets$ Eq. (\ref{eq:veracity});\\
        } 
        \algorithmiccomment{\textit{Multi-task Training}}\\
        Optimize $\mathcal{L}_{all} = \beta_D \mathcal{L}_D + \beta_S \mathcal{L}_S + \beta_C \mathcal{L}_C \gets $ Eq.  (\ref{eq:doc_loss},\ref{eq:sent_loss},\ref{eq:ver_loss});\\
        Update $\beta_D, \beta_S, \beta_C$;
\end{algorithm} \vspace{-0.2cm}

\renewcommand{\arraystretch}{0.7} 
\begin{table}[t!] 
\normalsize 
	\centering 

    \setlength{\tabcolsep}{1.2mm}{
		\begin{tabular}{l c c c} 
			\toprule
			\multicolumn{4}{c}{\textbf{RAWFC}} \\
			\midrule
			Annotator   &Gold &Exp-GenFE-MT &Exp-CofCED \\
			\midrule
			\multicolumn{4}{c}{$<$Informativeness$>$} \\
			\midrule
            \# 1         &\textbf{1.38}   &2.17    &\underline{1.89}       \\ 
            \# 2         &\textbf{1.63}   &2.32    &\underline{2.01}   \\
            \# 3        &\textbf{1.24}  &\underline{1.76}   &2.05      \\ 
            ALL        &\textbf{1.42}   &2.08    &\underline{1.98}		 \\
            \midrule
            
            \multicolumn{4}{c}{$<$Readability$>$} \\
			\midrule
            \# 1         &\textbf{1.74}   &1.98    &\underline{1.81}    \\ 
            \# 2         &\textbf{1.15}   &1.76    &\underline{1.63}   \\
            \# 3         &\textbf{1.97}   &2.35    &\underline{2.07}    \\ 
            ALL        &\textbf{1.62}   &2.03    &\underline{1.84}	   \\
            \midrule
            \multicolumn{4}{c}{$<$Overall$>$} \\
			\midrule
            \# 1         &\textbf{1.54}   &\underline{1.98}    &2.13		 \\
            \# 2         &\textbf{1.43}   &1.76    &\underline{1.73}   \\
            \# 3         &\textbf{1.60}   &2.24    &\underline{1.91}    \\ 
            ALL        &\textbf{1.52}   &1.99    &\underline{1.94}		 \\
            \midrule
            \midrule
            \multicolumn{4}{c}{\textbf{LIAR-RAW}} \\
            \midrule
            \multicolumn{4}{c}{$<$Informativeness$>$} \\
			\midrule
            \# 1         &\textbf{1.27}   &1.91    &\underline{1.82}    \\ 
            \# 2         &\textbf{1.55}   &2.09    &\underline{1.63}   \\
            \# 3         &\textbf{1.12}   &1.72    &\underline{1.46}    \\ 
            ALL          &\textbf{1.31}   &1.91    &\underline{1.64}	   \\
            \midrule

			\multicolumn{4}{c}{$<$Readability$>$} \\
			\midrule
            \# 1         &\textbf{1.13}   &2.29    &\underline{1.78}       \\ 
            \# 2         &\textbf{1.38}   &2.25    &\underline{2.12}   \\
            \# 3         &\textbf{1.24}   &\underline{1.94}    &2.02      \\ 
            ALL          &\textbf{1.25}    &2.16    &\underline{1.97}		 \\
            \midrule
            
            \multicolumn{4}{c}{$<$Overall$>$} \\
			\midrule
            \# 1         &\textbf{1.33}   &1.96    &\underline{1.68}    \\ 
            \# 2         &\textbf{1.49}   &2.12    &\underline{1.94}   \\
            \# 3         &\textbf{1.51}   &2.35    &\underline{2.08}    \\ 
            ALL        &\textbf{1.44}   &2.14    &\underline{1.90}		 \\
            \bottomrule
		\end{tabular} }
	\caption{
	Mean Average Ranks (MAR) of the explanations for each three evaluation criteria on RAWFC and LIAR-RAW, respectively.  Gold denotes the explanations come from the justification, Exp-GenFE-MT denotes the explanations generated by GenFE-MT, and Exp-CofCED denotes the explanations generated by our CofCED. Best performances are shown in bold, and the second ones are underlined. 
	} 
	\label{tab:mar} \vspace{-0.4cm}
\end{table} 
\section{Human Evaluation for Explanations}
\label{app:human}
We also study the explanation quality by human evaluation referring to \cite{atanasova2020generating}. Provided with three types of explanations, i.e., human justification, veracity explanation generated by CofCED, and the ones generated by GenFE-MT, three English-speaking adult annotators were asked to rank them with 1--Good, 2--Medium, 3--Poor, according to three different criteria. To keep clear and simple, we use the following criteria: 

\begin{itemize}
    \item \textbf{Informativeness}. The explanation contains much evidential information that contributes to fake news detection. 
    \item \textbf{Readability}. The explanation is easy to understand. 
    \item \textbf{Overall}. The explanation is ranked based on their overall quality. 
\end{itemize}
For the annotation settings, we randomly sample a set of 40 instances from the test set and prepare three candidate explanations without any other information about these explanations. All of annotators work independently.

Table \ref{tab:mar} shows the mean average results from the manual evaluation. 
We also compute Krippendorff’s inter-annotator agreement~\cite{atanasova2020generating} and obtain 0.37 for Informativeness, 0.43 for Readability, 0.31 for Overall. 
From the results, we can see that the human justification (Gold) achieves the best quality and our Exp-CofCED achieves better quality of explanations than Exp-GenFE-MT. 
These results suggest that the ROUGE results in Table~\ref{tab:exp-rouge} may be not sufficient for evaluating veracity explanations because the ROUGE score only accounts for word overlapping. 
Besides, the performance of veracity prediction in Table~\ref{tab:veracity-acc} also verifies the effectiveness of explanations in improving fake news detection. In summary, our proposed CofCED can 
significantly improve final fake news detection with overall better veracity explanations. 

\section{Further Discussion} 
\setcounter{subsection}{0}
\setcounter{equation}{0}
\setcounter{table}{0}
\setcounter{figure}{0}
\renewcommand{\thesubsection}{E.\arabic{subsection}}
\renewcommand{\theequation}{E.\arabic{equation}}
\renewcommand{\thetable}{E.\arabic{table}}
\renewcommand{\thefigure}{E.\arabic{figure}}

\renewcommand{\arraystretch}{0.7} 
\begin{table}[t!] 
\normalsize 
	\centering 

    \setlength{\tabcolsep}{2.0mm}{
		\begin{tabular}{l c c c} 
			\toprule
			
			Dataset   & P(\%)&R(\%)&macF1(\%) \\
			\midrule
			RAWFC &84.28  &79.29  &81.71 \\
            LIAR-RAW &14.98    &61.06    &24.06  		 \\
			\bottomrule 
		\end{tabular} }
	\caption{
	Our results on report classification.  
	} 
	\label{tab:exp-doc} \vspace{-0.4cm}
\end{table} 
Table \ref{tab:exp-doc} shows internal results about report classifications regarding precision, recall, and macro F1 score. Our model outperforms better on RAWFC than on LIAR-RAW, indicating that report classification for fine-grained claims is much challenging and further improving this part may contribute to explainable fake news detection.  Similarly, Table~\ref{tab:exp-sent} shows internal results about explainable sentence classifications. Overall, our CofCED significantly outperforms GenFE-MT but only achieves comparable results on LIAR-RAW in terms of ROUGE scores (Table~ \ref{tab:exp-rouge}). This is probably because ROUGE scores w.r.t. word overlapping are not sufficient for evaluating the qualities of generated explanations. Thus, we further introduce human evaluation as a complementary measure. 

\renewcommand{\arraystretch}{0.85} 
\begin{table*}[t!] 
\normalsize 
	\centering 
    \setlength{\tabcolsep}{2.0mm}{
		\begin{tabular}{l c c c   c c c} 
			\toprule
			
			\multirow{2}{*}{Model}& 
			\multicolumn{3}{c}{RAWFC} &\multicolumn{3}{c}{LIAR-RAW}\\
			\cmidrule(lr){2-4} \cmidrule(lr){5-7}  &P(\%)&R(\%)&macF1(\%) &P(\%)&R(\%)&macF1(\%)\\
			\midrule
			GenFE-MT \cite{atanasova2020generating} &50.62 &36.03 &42.09 	        &\textbf{43.83}  &4.27  &7.79 \\ 
          CofCED                        &\textbf{55.56}  &\textbf{41.67}  &\textbf{47.62}     &14.29  &\textbf{22.22}  &\textbf{17.39}   		 \\
			\bottomrule 
		\end{tabular} }
	\caption{
	Experimental results of explainable sentence classification regarding oracle sentences. 
	} 
	\label{tab:exp-sent} \vspace{-0.2cm}
\end{table*} 

\section{Example}
\setcounter{subsection}{0}
\setcounter{equation}{0}
\setcounter{table}{0}
\setcounter{figure}{0}
\renewcommand{\thesubsection}{F.\arabic{subsection}}
\renewcommand{\theequation}{F.\arabic{equation}}
\renewcommand{\thetable}{F.\arabic{table}}
\renewcommand{\thefigure}{F.\arabic{figure}}

\label{sec:example}
Examples from RAWFC are shown in Table \ref{example}.

\begin{table*}[!htbp]
    \centering
    \begin{tabular}{p{1.0\linewidth}}
     \toprule 
     $[$\textbf{Label:} False$]$ \textbf{Claim:} \textit{U.S. Rep. Alexandria Ocasio-Cortez started “chain migration" deportation proceedings against First Lady Melania Trump and her parents.}\\
     \textbf{Explanation:} Illegal immigration remained a top issue for U.S. President Donald Trump and continued to divide Americans in mid-2019, all the more so after Trump told several Democratic members of Congress of immigrant parentage, all but one of them born in the United States, they should “go back and help fix the totally broken and crime infested places from which they came.” (...) 
     This is simply not true. For context, “chain migration” is a term used to describe immigration procedures that allow adult U.S. citizens to obtain citizenship for foreign-born adult relatives. Reportedly, the first lady’s parents secured their citizenship through just such a procedure — though we needn’t belabor the point, because everything else in the story is fictional (Melania Trump’s parents aren’t named “Oedipus and Jezebel Beelzebub.'' 
    \\
    \textbf{Raw Report Domain:} \textit{www.newsweek.com}\\
    \textbf{Content:} The president have also be criticize for want to end ``chain migration'',  a program that let U.S. citizen to sponsor immediate family member for legal residency, despite it be the program that Melania Trump use to put her parent Viktor and Amalija Knavs on a path to American citizenship. (...)\\
    \textbf{Raw Report Domain:} \textit{www.washingtonpost.com}\\
    \textbf{Content:} Melania Trump' s parent be legal permanent resident, raise question about whether they rely on ``chain migration'' She enjoy put her personal mark on the historic home and have redesign the family live quarter. (...) \\
    \textbf{Raw Report Domain:} \textit{www.kbzk.com}\\
    \textbf{Content:} Melania Trump’s parent, Viktor and Amalija Knavs, also go through the immigration process, use the perjoratively call “chain migration” route the President have criticize. (...) A source with direct knowledge of Melania Trump’s parent and their immigration status previously tell CNN that she have sponsor her parent for their green card, a status that allow them to live and work in the US indefinitely and pave the way for citizenship. (...)\\
    \midrule \midrule 
     $[$\textbf{Label:} True$]$ \textbf{Claim:} \textit{The snakehead fish can survive on land.} \\
     \textbf{Explanation:} On Oct.10, 2019, many readers came across news stories about an invasive species of fish called the snakehead fish that had been discovered in Georgia. While these stories largely dealt with wildlife officials’ attempts to eradicate the species, what caught the attention of most readers were brief mentions of this fish’s unique ability to survive on land.  CNN reported:  A snakehead fish that survives on land was discovered in Georgia. Officials want it dead  An invasive fish species that can breathe air and survive on land has been found in Georgia for the first time. And officials are warning anyone who comes into contact with the species to kill it immediately.
     The snakehead fish can truly survive on land. Here’s a video of a snakehead in Thailand as it “walks,” crawls, or wiggles its way back to the water. \\
    \textbf{Raw Report Domain:} \textit{ www.cbsnews.com}\\
    \textbf{Content:} Northern snakehead be invasive fish that can breathe air and survive for day on land. Lawrenceville, Georgia — Georgia's Department of Natural Resources have a message for angler: If you catch a northern snakehead, kill it immediately. \\
    \textbf{Raw Report Domain:} \textit{www.nytimes.com}\\
    \textbf{Content:} Snakeheads can survive in freshwater and be describe a predator that can eat tiny animal, and travel across land, live out of water for several day. There have be no end to the creepy description of the snakehead fish, a slimy, toothy, large-jawed animal that can breathe on land and crawl like a snake, in the decade that it have pop up in freshwater lake, pond and river in the United States. (...)\\
    \bottomrule  
     \end{tabular}
     \caption{Examples from RAWFC. } 
     \label{example}
\end{table*}


\end{document}